\documentclass[letterpaper]{article} 
\usepackage{aaai25}  
\usepackage{times}  
\usepackage{helvet}  
\usepackage{courier}  
\usepackage[hyphens]{url}  
\usepackage{graphicx} 
\usepackage{natbib}  
\usepackage{caption} 
\usepackage{algorithm}
\usepackage{algorithmic}

\usepackage{newfloat}
\usepackage{listings}
\DeclareCaptionStyle{ruled}{labelfont=normalfont,labelsep=colon,strut=off} 
\floatstyle{ruled}
\newfloat{listing}{tb}{lst}{}
\floatname{listing}{Listing}
\title{Assessing Pre-Trained Models for Transfer Learning Through \\Distribution of Spectral Components}

\author{
    Tengxue Zhang\textsuperscript{\rm 1},
    Yang Shu\textsuperscript{\rm 1}\thanks{Corresponding author},
    Xinyang Chen\textsuperscript{\rm 2}$\footnotemark[1]~$, 
    Yifei Long\textsuperscript{\rm 1},
    Chenjuan Guo\textsuperscript{\rm 1},
    Bin Yang\textsuperscript{\rm 1}
}
\affiliations{

    \textsuperscript{\rm 1}School of Data Science and Engineering, East China Normal University\\
    \textsuperscript{\rm 2}School of Computer Science and Technology, Harbin Institute of Technology (Shenzhen)\\
    \{txzhang, yflong\}@stu.ecnu.edu.cn,
    \{yshu, cjguo, byang\}@dase.ecnu.edu.cn, chenxinyang@hit.edu.cn

}

\usepackage{bibentry}
\usepackage{subcaption}
\usepackage{adjustbox}
\usepackage{booktabs}
\usepackage{amsmath}
\usepackage{amssymb}
\usepackage{amsfonts}
\usepackage{bm}
\usepackage{multirow}

\begin{document}

\maketitle

\begin{abstract}
Pre-trained model assessment for transfer learning aims to identify the optimal candidate for the downstream tasks from a model hub, without the need of time-consuming fine-tuning. Existing advanced works mainly focus on analyzing the intrinsic characteristics of the entire features extracted by each pre-trained model or how well such features fit the target labels. 
This paper proposes a novel perspective for pre-trained model assessment through the Distribution of Spectral Components (DISCO). Through singular value decomposition of features extracted from pre-trained models, we investigate different spectral components and observe that they possess distinct transferability, contributing diversely to the fine-tuning performance.
Inspired by this, we propose an assessment method based on the distribution of spectral components which measures the proportions of their corresponding singular values. Pre-trained models with features concentrating on more transferable components are regarded as better choices for transfer learning. We further leverage the labels of downstream data to better estimate the transferability of each spectral component and derive the final assessment criterion. Our proposed method is flexible and can be applied to both classification and regression tasks. We conducted comprehensive experiments across three benchmarks and two tasks including image classification and object detection, demonstrating that our method achieves state-of-the-art performance in choosing proper pre-trained models from the model hub for transfer learning.
\end{abstract}

%

\section{Introduction}
\begin{figure}[ht]
    \centering
    \includegraphics[width=0.47\textwidth]{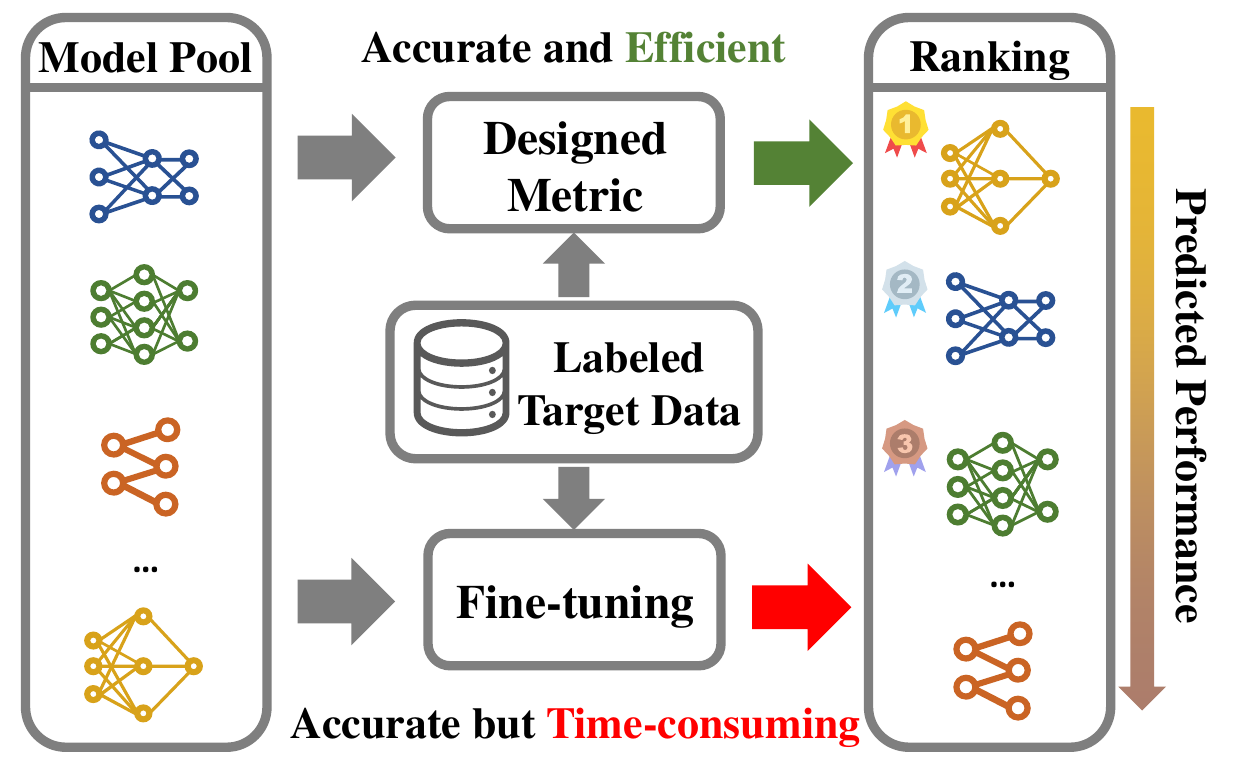}
    \caption{The overall framework for assessing pre-trained models for transfer learning. Given a model pool, a designed metric quickly predicts and ranks model performance on the target dataset. The predicted rankings are expected to strongly correlate with the ground-truth fine-tuning results.}
    \label{fig:intro}
\end{figure}

With the widespread use of deep neural networks, numerous pre-trained models are readily available in the artificial intelligence community, such as those found on HuggingFace \cite{wolf2019huggingface} and TensorFlowHub \cite{abadi2016tensorflow}. 
Thanks to being trained on large-scale datasets, these models provide fundamental and general knowledge for downstream tasks. Fine-tuning is a successful paradigm that leverages this knowledge to enhance performance on target tasks, without having to start the training process from scratch. 
Despite the remarkable capabilities of pre-trained models, such as those pre-trained on large-scale images for computer vision~\cite{russakovsky2015imagenet,he2016ResNet,dosovitskiy2020vit}, and large corpus for natural language processing~\cite{devlin2018bert,brown2020languagegpt3}, there is no single pre-trained model that excels in all downstream tasks across various domains. Consequently, selecting the best pre-trained model to apply to downstream tasks has become a challenging problem.
Fine-tuning each pre-trained model on the target dataset is a straightforward way to obtain the ground-truth performance and select the best model. However, this procedure can be time-consuming with a large-scale pool of pre-trained models. As a result, recent works have focused on designing a metric to quickly predict the performance of models on the target dataset for pre-trained model selection \cite{ding2024which}, as illustrated in Figure \ref{fig:intro}. 

Most of these works calculate the transferability metric based on the original features of the target dataset extracted by pre-trained models. Some primarily consider statistical information \cite{nguyen2020LEEP}, such as the expected conditional probability of target labels given the pre-training labels, to evaluate how well the extracted features fit the target labels. Others focus on the intrinsic characteristics of the entire extracted features, such as inter-class separation or intra-class and inter-class structure \cite{wang2023NCTI}.
 
One significant limitation of previous methods is their reliance on the original extracted features, often overlooking the fine-grained changes occurring during the fine-tuning process. For cases where the difference between the source and target domains is substantial, the extracted features obtained from fine-tuned models may differ considerably from those before fine-tuning. In such situations, a transferability metric based solely on the original extracted features of pre-trained models and the fitness between such features and the target labels can be unreliable.
Despite some works considering the dynamics in the fine-tuning process, their metrics are designed based on hypotheses specific to class relationships, making them primarily only applicable to classification tasks. For example, the class separability hypothesis of the fine-tuning process is explored in methods like GBC \cite{pandy2022GBC}, which directly calculates the pair-wise class overlap using the Bhattacharyya coefficients and SFDA \cite{shao2022SFDA}, which projects the extracted features into a more class-separable Fisher space. 

In this work, we consider the fine-grained changes in extracted features from a novel perspective of the \textbf{DI}stribution of \textbf{S}pectral \textbf{CO}mponents (\textbf{DISCO}) and propose a flexible framework for assessing models for transfer learning. First, we resort to singular value decomposition (SVD) to investigate the changes in the extracted features before and after fine-tuning. We observe that, during the fine-tuning process, different spectral components are under varying degrees of change and exhibit distinct transferability. After fine-tuning, the distribution of singular values tends to become more concentrated on more transferable components. 
Leveraging these observations and considering the potential effect of different spectral components on the fine-tuning process, we design a framework to evaluate pre-trained models through the distribution of the spectral components.
Furthermore, taking into account the label information of the downstream task, we design different metrics based on this framework for two general task types: classification and regression tasks. The successful implementation in both the image classification and the object detection tasks demonstrates that our method is more flexible and can be customized with specific metrics for different downstream tasks.
In summary, the major contributions of our work are as follows:
\begin{itemize}
\item[$\bullet$] We investigate the fine-grained changes in extracted features during the fine-tuning process through the perspective of spectral components. We propose an assessment framework for pre-trained models based on the distribution of spectral components.
\item[$\bullet$] Our proposed framework is highly flexible. We design assessment metrics tailored to the two general tasks: classification and regression. These metrics are then successfully applied in image classification and object detection.

\item[$\bullet$]  
Extensive experiments on three benchmarks and two specific tasks, demonstrate that our method achieves state-of-the-art performance in assessing pre-trained models for transfer learning.
\end{itemize}

\section{Related Work}
\paragraph{Transfer Learning}
Transfer learning leverages transferring knowledge from the source domains to enhance performance in the target domain. One prominent branch of transfer learning is inductive transfer, commonly known as fine-tuning \cite{erhan2010unsupre-train, yosinski2014transferable}, which utilizes pre-trained models to improve target task performance, especially when labeled data is limited. Various strategies exist for fine-tuning, ranging from Full Fine-Tuning (FFT) \cite{donahue2014decaf, radford2021learning} to Parameter-Efficient Fine-Tuning (PEFT) \cite{kumar2022LPFT, jia2022visual, liu2023gpt, dettmers2024qlora}. 
However, this process can be time-consuming due to the iterative nature of training and the need to select appropriate hyperparameters. Despite the availability of numerous pre-trained models, choosing the most effective pre-trained model for a specific target task without requiring a substantial time investment remains a significant challenge.

\paragraph{Pre-trained Model Selection}
Pre-trained model selection seeks to quickly identify the best model for downstream tasks from a model zoo without extensive fine-tuning. Previous works like NCE  \cite{tran2019NCE} and LEEP \cite{nguyen2020LEEP} have focused on probabilistic methods based on the expected empirical distribution of the target labels. However, these metrics are not suitable for self-supervised pre-trained models due to their reliance on pre-trained classifiers. 
To address this limitation, LogME \cite{you2021LogME} estimates maximum label marginalized likelihood, and NLEEP \cite{nguyen2020LEEP} extends LEEP by replacing the output layer with a Gaussian mixture model. GBC \cite{pandy2022GBC} calculates the overlap between pairwise target classes in extracted features. SFDA improves class separability using a Fisher space and then applies a Bayesian classifier. PED \cite{li2023PED} stimulates dynamics during fine-tuning through the lens of potential energy and integrates into existing model selection metrics.
TMI \cite{xu2023TMI} uses intra-class feature variance as a performance indicator, positing that lower variance reflects tighter class feature clustering.
Inspired by neural collapse phenomena, NCTI \cite{wang2023NCTI} develops metrics to measure the distance from the current status of pre-trained models to their hypothetical fine-tuned state. 
Different from these methods that primarily rely on the entire extracted features, we propose a novel perspective for pre-trained model selection through the distribution of spectral components.

\section{Method}
\label{headings}
\begin{figure*}[ht]
    \centering
    \includegraphics[width=\textwidth]{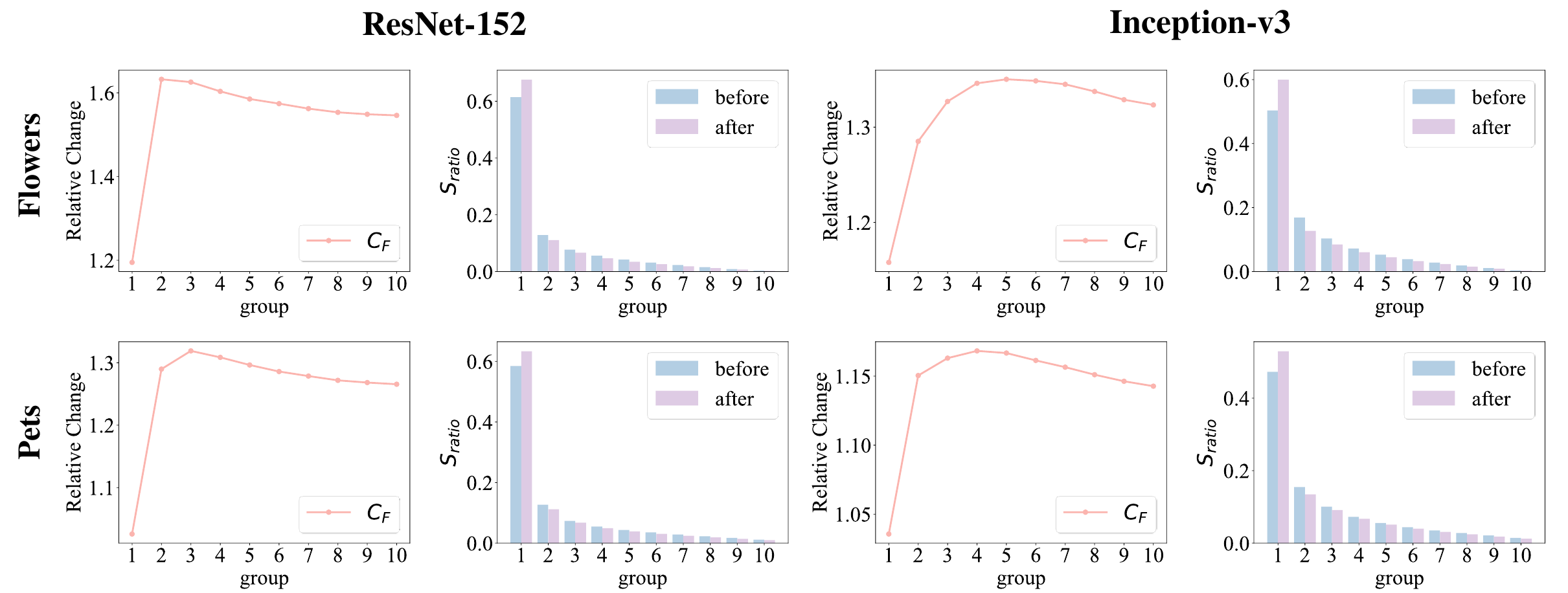}
    \caption{The relative changes of Frobenius norm $C_F$ and the proportion of singular values $S_{\text{ratio}}$ in different spectral components of extracted features before and after fine-tuning.}
    \label{fig:all_figure}
\end{figure*}
\subsection{Problem Formulation}
\label{problem formulation}
Given a pool of $M$ pre-trained models $\{\phi^m\}_{m=1}^M$ and a target dataset $D = \{x_i, y_i\}_{i=1}^{N}$ with $N$ samples, pre-trained model selection aims to efficiently assess and select the optimal model for transfer learning. Typically, in the fine-tuning paradigm, we initialize a predictor head on the backbone of the pre-trained model and then fine-tune the entire network on the target dataset. By fine-tuning each model in a brute-force manner, we obtain the ground-truth performance $\{P^m\}_{m=1}^M$ for the model hub. To avoid this time-consuming process, model selection methods compute an assessment score $S^m$ for each pre-trained model $\phi^m$. A higher $S^m$ suggests that the model $\phi^m$ is likely to perform well when transferred to the target dataset.
Ideally, the calculated scores $\{S^m\}_{m=1}^M$ for the model hub should strongly correlate with their actual fine-tuning performance $\{P^m\}_{m=1}^M$, allowing us to identify the best model for transfer.


\subsection{Transferability Assessment Framework through Spectral Component Distribution}
\label{sec:empirical}

\begin{figure*}[ht]
    \centering
    \includegraphics[width=\textwidth]{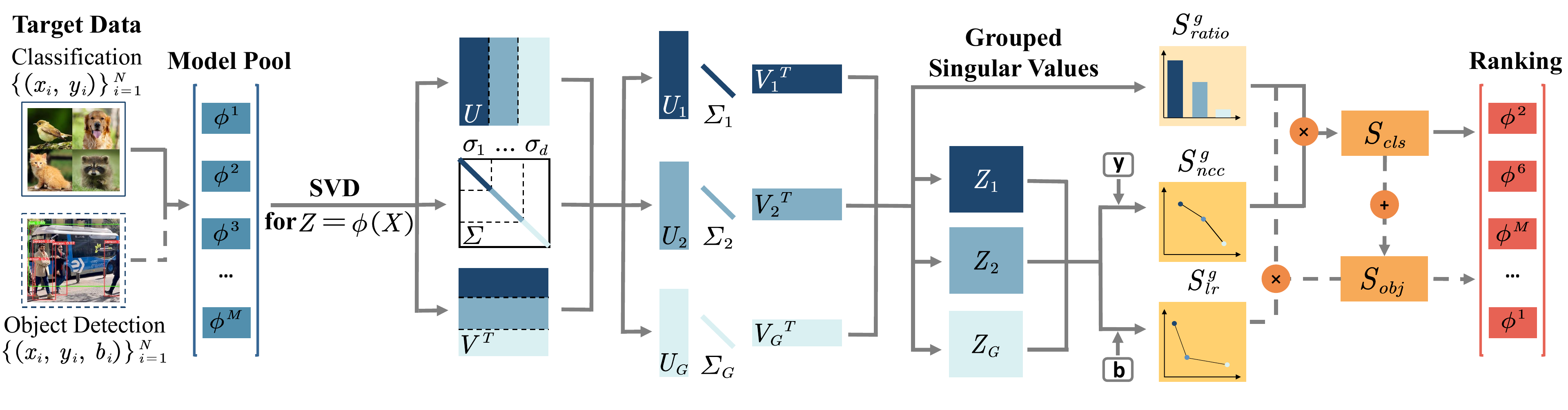}
    \caption{Overview of DISCO's framework (better viewed in color). $S_\text{ratio}^g$ represents the singular value ratio of the $g$-th spectral component, while $S_{\text{ncc}}^g$ and $S_{\text{lr}}^g$ are task-specific scores for classification and regression tasks, respectively. The overall transferability of the entire feature is calculated through the perspective of the distribution of spectral components.}
    \label{fig:model}
\end{figure*}
To explore fine-grained changes in extracted features during fine-tuning, we employ singular value decomposition (SVD) for in-depth analysis. We fine-tune ResNet-152 \cite{he2016ResNet}, DenseNet-201 \cite{huang2017DenseNet}, and Inception-v3 \cite{szegedy2016Inceptionv3} on Caltech101 \cite{fei2004Caltech}, Oxford-102 Flowers \cite{nilsback2008Flowers}, and Pets \cite{parkhi2012Pets} respectively. Let $\phi^m$ represent the $m$-th pre-trained model and $\phi'^m$ the fine-tuned model. The features extracted by $\phi^m$ and $\phi'^m$ are denoted as $\mathbf{Z}\in \mathbb{R}^{N \times d}$ and $\mathbf{Z'}\in \mathbb{R}^{N \times d} $ respectively, where $N$ is the target dataset size and $d$ is the feature dimension. We apply SVD to extracted features $\mathbf{Z}$ and $\mathbf{Z'}$ as follow:
\begin{equation}
    \mathbf{Z} = \mathbf{U}\mathbf{\Sigma}\mathbf{V}^{\mathsf{T}}, \quad \mathbf{Z'}= \mathbf{U'}\mathbf{\Sigma'} \mathbf{V}'^{\mathsf{T}}
\end{equation}
where $\mathbf{U}$ and $\mathbf{U'}$ are orthogonal matrices of left singular vectors,  while $\mathbf{V}$ and $\mathbf{V'}$ are orthogonal matrices of right singular vectors, $\mathbf{\Sigma}$ and $\mathbf{\Sigma'} $ are diagonal matrices of non-negative singular values arranged in descending order. 
We divide the original feature $\mathbf{Z}$ into different spectral components by grouping the singular values in descending order into $G$ groups, each containing $s$ consecutive singular values. Similar operations are performed on $\mathbf{U}$ and $\mathbf{V}$.  The $g$-th spectral component is defined as the product of the matrices corresponding to the $g$-th group's columns of $\mathbf{U}_g$, the diagonal singular values $\mathbf{\Sigma}_g$, and rows of $\mathbf{V}_g^\mathsf{T}$. 
In this way, we obtain different spectral components $\mathbf{Z}_g$ and $\mathbf{Z'}_g$ of the original feature $\mathbf{Z}$ as truncated SVD reconstruction. We perform the same process for $\mathbf{Z'}$ for consistency:
\begin{equation}
    \mathbf{Z}_g = \mathbf{U}_g\mathbf{\Sigma}_g \mathbf{V}^\mathsf{T}_g, \quad 
    \mathbf{Z}_g' = \mathbf{U}_g'\mathbf{\Sigma}_g'\mathbf{V}_g'^{\mathsf{T}}
\end{equation}

We investigate the changes in different spectral components of the extracted features before and after fine-tuning from two perspectives. The first perspective is to compute the relative change of different spectral components using the Frobenius norm, which can be calculated as: 

\begin{equation}
    C_F^g = \frac{\|\mathbf{Z}_g-\mathbf{Z'}_g\|_F}{\|\mathbf{Z}_g\|_F}
\end{equation}
We set $G=10$ in our experiments. Partial results are shown in Figure \ref{fig:all_figure}, with the complete results available in Appendix
\ref{app:entireemp}.
We observe that the spectral component with the larger singular value generally undergoes the least variation while variation increases in subsequent groups and then decreases slightly for components with smaller singular values. These differences suggest varying levels of transferability during fine-tuning, with the components with larger singular values often exhibiting higher transferability. 

Another perspective is to compute the singular value ratio $S_{\text{ratio}}^g$ for singular values $\sigma^g$ of each spectral component:
\begin{equation}
\label{score:group}
    S_{\text{ratio}}^g = \frac{\sum_{j=1}^s\sigma_j^g}{\sum_{i=1}^d \sigma_i}
\end{equation}
where $\sigma_i$ is the $i$-th singular value in $\mathbf{\Sigma}$. By analyzing the proportional alterations of singular values across different spectral components, we observe that after fine-tuning, the larger singular values increase in proportion, while smaller ones diminish. This suggests that the distribution of singular values tends to concentrate on components with larger singular values during fine-tuning, indicating that more transferable components play a greater role in fine-tuned features.

Inspired by our empirical observations of the fine-grained changes during fine-tuning, the distribution of spectral components influences the transfer learning process and results. We first propose an intuitive solution to assess a model's transferability. The key insight is that a concentration of singular values on more transferable spectral components suggests greater overall transferability of entire features. To quantify this, we calculate the ratio of top-$k$ groups' singular values to all singular values for each pre-trained model:
\begin{equation}
\label{score:topk}
    S_{\text{topk}} = \frac{\sum_{g=1}^k\sum_{j}^s \sigma_j^g}{\sum_{i=1}^d\sigma_i}
\end{equation}
However, in practice, the distribution of singular values varies across different models, making it difficult to draw a hard boundary of $k$, where these top-$k$ components are considered transferable and the rest are deemed not. Focusing solely on the top-$k$ components ignores the potential contributions of other ones. Moreover, without task-specific evaluation, the actual performances of different spectral components on target tasks are uncertain, particularly in cases with significant domain gaps. Therefore, to quantitatively assess the transferability and enable comparisons between different pre-trained models, it is crucial to leverage the labels from the target dataset to obtain the performance on downstream tasks. 
Based on these considerations, we propose our DISCO framework that quantifies the overall transferability of each pre-trained model through the perspective of the distribution of spectral components in their extracted features: 
\begin{equation}
\label{score:overall}
    S_{\text{DISCO}} = \sum_{g=1}^G \ S_{\text{task}}^g * S_{\text{ratio}}^g
\end{equation}
Here, $S_{\text{task}}^g$ represents the quantitative transferability of the $g$-th spectral component on specific downstream tasks, while $S_{\text{ratio}}^g$ is its corresponding singular value ratio. $S_{\text{task}}^g$ captures how well this component performs when transferred to new tasks, and $S_{\text{ratio}}^g$ reflects the importance of this component within the entire feature. The summation considers each spectral component's contribution to the overall transferability of the entire extracted feature.
Pre-trained models with features concentrated in more transferable components, and with such components demonstrating superior downstream performance, are considered more suitable candidates for transfer learning.


\subsection{Task-Specific Performance Assessment for Spectral Components}
\label{sec:tasks-spcific}

Classification and regression are foundational tasks in machine learning, serving as the basis for many specific applications. In this section, we extend our framework by leveraging target labels and designing specific metrics for $S_{\text{task}}^g$. The overview of our framework is shown in Figure \ref{fig:model}. By tailoring to both classification and regression tasks, our framework is flexible and applicable to various downstream tasks, including image classification and object detection.


\subsubsection{Classification Score}
Due to the domain gap, pre-trained classifiers from the source domain are not directly applicable to the target task. Therefore, the most straightforward way to evaluate the performance of different spectral components is by assessing their ability to discriminate between classes. Under a commonly used assumption that the class distribution follows a multivariate Gaussian distribution, we use the performance of the nearest centroid classifier as a measure. Given a sample $x_i$ with corresponding feature $\mathbf{z}_i^g$ from the $g$-th spectral component, the posterior probability that it belongs to the $c$-th class using Mahalanobis distance can be calculated as:
\begin{equation}
P(y=c | \mathbf{z}_i^g) = -\frac{1}{2}(\mathbf{z}_i^g-\bm{\mu}_c)^\mathsf{T}\mathbf{\Lambda}_c^{-1}(\mathbf{z}_i^g-\bm{\mu}_c) + \log(\pi_c)
\end{equation}
where $\mathbf{\Lambda}_c$ and $\bm{\mu}_c$ represent the covariance matrix and mean of the $c$-th class. The prior probability $\pi_c$ is given by $N_c/N$, where $N_c$ is the number of the samples in class $c$, and $N$ is the total number of samples.
We apply the softmax function to normalize the posterior probability of the $i$-th sample $\mathbf{z}_i^g$ predicted by the $g$-th spectral component as:
\begin{equation}
    \hat{y}_{i,c}^g = \frac{\text{exp}(\log P(y=c|\mathbf{z}_i^g))}{\sum_{k=1}^C\text{exp}(\log P(y=k|\mathbf{z}_i^g))}
\end{equation}
where $\hat{y}_{i,c}^g$ is the final prediction probability that the $i$-th sample belongs to class $c$.
As such, we quantify the average confidence of the nearest centroid classifier on
the $g$-th spectral component as the score for classification performance:
\begin{equation}
    S_{\text{ncc}}^g = \frac{1}{N}\sum_{i=1}^N \mathbf{y}_i^\mathsf{T}\mathbf{\hat{y}}_i^g
\end{equation}
where $\mathbf{y}_i$ denotes the one-hot ground truth label of $x_i$ and $\mathbf{\hat{y}}_i^g$ is the $C$ dimensional logits predicted by the $g$-th spectral component, with each dimension representing one class. 

For image classification, the goal is to correctly classify images. Therefore, we use $S_{\text{ncc}}^g$ as the performance assessment for the $g$-th spectral component. Based on the proposed framework, we derive the final assessment criterion for each pre-trained model in image classification as follows:
\begin{equation}
    S_{\text{cls}} = \sum_{g=1}^G S_{\text{ncc}}^g *S_{\text{ratio}}^g  
\end{equation}
\subsubsection{Regression Score}
Regression is a task for predicting continuous numerical values. To stimulate the target task fine-tuning and avoid significant time expenditure, we propose a linear approximation between extracted features and their labels. 
In linear regression, our objective is to find the coefficients $\mathbf{\beta}$ to minimize ordinary least-square (OLS) distance $\|y-\mathbf{Z}\mathbf{\beta}\|^2$.
However, directly solving this may be problematic due to the potential non-invertibility of $\mathbf{Z}$. Singular value decomposition (SVD) offers a robust and numerically stable method for matrix inversion. Since SVD is used to divide spectral components, it adds no extra time cost for the regression score. Therefore, we employ SVD to decompose the extracted features $\mathbf{Z}$ and compute its pseudo-inverse $\mathbf{Z}^\dagger$ as:
\begin{equation}
    \mathbf{Z} = \mathbf{U}\mathbf{\Sigma}\mathbf{V}^\mathsf{T}, \quad
    \mathbf{Z}^\dagger = \mathbf{V}\mathbf{\Sigma}^{-1}\mathbf{U}^\mathsf{T}
\end{equation}
where $\mathbf{U}$ and $\mathbf{V}$ are orthogonal matrices, and $\mathbf{\Sigma}$ is a diagonal matrix containing the singular values of $\mathbf{Z}$.
The coefficient $\beta$ can be calculated using the pseudo-inverse: $\beta = \mathbf{Z}^\dagger y$. 


In the object detection task, where the goal is to predict bounding box coordinates $b$ along with category labels $y$, we calculate the approximated bounding box coordinates as $\hat{b} = \mathbf{Z}\hat{\mathbf{Z}}^\dagger b$. 
To evaluate the quality of these approximations, we employ the mean squared error (MSE) between the ground truth and approximated bounding box labels for the $g$-th spectral component. Given a total of $K$ bounding boxes in the target dataset, the regression score $S_{\text{lr}}^g$ for the $g$-th spectral component is defined as: 
\begin{equation}
    S_{\text{lr}}^g = -\frac{1}{K \times 4} \sum_{k=1}^{K}\sum_{j=1}^{4}(b_{k}^{(j)}- \hat{b}_{k}^{(j)})^2
\end{equation}
where $b_{k}^{(j)}$ and $\hat{b}_{k}^{(j)}$ are the $j$-th ground-truth and approximated coordinate of the $k$-th bounding boxes, respectively.
Similarly, based on the proposed framework, we use $S_{\text{lr}}^g$ to assess the performance of the $g$-th spectral component. The final assessment for each pre-trained model in the regression task is as follows:
\begin{equation}
    S_{\text{reg}} = \sum_{g=1}^G S_{\text{lr}}^g * S_{\text{ratio}}^g 
\end{equation}
Since both classification and regression scores are required in object detection and they have different scales, we use min-max normalization to standardize each score to a $[0, 1]$ range, avoiding the need for weighted adjustments.
We sum the normalized $S_{\text{cls}}$ and $S_{\text{reg}}$ to obtain the final model selection score for each pre-trained models for object detection: 
\begin{equation}
    S_{\text{obj}} = S_{\text{cls}} + S_{\text{reg}}
\end{equation}

\begin{table*}

  \setlength{\tabcolsep}{1mm} 
  \centering
  \begin{tabular}{lcccccccccccc}
    \toprule
& Aircraft &Caltech &Cars &Cifar10 &Cifar100 &DTD &Flowers &Food &Pets &SUN &VOC &Average \\
    \midrule
NCE \shortcite{tran2019NCE}& -0.161 & 0.465 & \underline{0.685} & 0.709 & 0.723 & 0.302 & -0.482 &0.627 & 0.772 & 0.760 & 0.571 & 0.452 \\
LEEP \shortcite{nguyen2020LEEP} & -0.277 & 0.605 & 0.367 & 0.824 & 0.677 & 0.486 & -0.291 & 0.434 & 0.389 & 0.658 & 0.413 & 0.390 \\
LogME \shortcite{you2021LogME}  & 0.439 & 0.463 & 0.605 & 0.852 & 0.725 & 0.700 & 0.147 & 0.385 & 0.411 & 0.511 & 0.695 &0.539\\
NLEEP \shortcite{li2021Nleep} & -0.531 & 0.614 & 0.489 & 0.825 & 0.731 & \textbf{0.820} & 0.054 & 0.529 & \textbf{0.955} & \textbf{0.848} & 0.699 & 0.548 \\
SFDA \shortcite{shao2022SFDA}  & \underline{0.614} & \textbf{0.696} & 0.518 & \textbf{0.949} & 0.866 & 0.575 & 0.514 & \underline{0.815} & 0.522 & 0.558 & 0.671 & 0.663 \\
Etran \shortcite{gholami2023etran}  & -0.091 & 0.440 & 0.246 & \underline{0.887} & \textbf{0.900} & 0.303 & \underline{0.580} & \textbf{0.829} & 0.713 & 0.708 & 0.667 & 0.562 \\
NCTI \shortcite{wang2023NCTI}  & 0.496 & \underline{0.679} & 0.647 & 0.843 & 0.879 & 0.704 & 0.541 & 0.773 & \underline{0.924} & 0.756 & \underline{0.741} & \underline{0.726} \\
 \midrule
DISCO & \textbf{0.652} & 0.661 & \textbf{0.795} & 0.823 & \underline{0.895} & \underline{0.764} & \textbf{0.712} & 0.678 & 0.575 & \underline{0.773} & \textbf{0.802} & \textbf{0.739} \\
    \bottomrule
  \end{tabular}
    \caption{Method comparison of their correlation strength with ground-truth fine-tuning accuracy on supervised models. Weighted Kendall's $\tau$ on 11 target datasets and their average are listed above. For each column, the best, and second-best results are in bold, and underlined, respectively. Our method achieves the best overall weighted Kendall’s $\tau_\omega$. }
  \label{table1}
\end{table*}
\begin{table*}
  \centering
  \setlength{\tabcolsep}{1mm} 
  \begin{tabular}{lcccccccccccc}
    \toprule
& Aircraft &Caltech &Cars &Cifar10 &Cifar100 &DTD &Flowers &Food &Pets &SUN &VOC &Average \\
    \midrule
LogME \shortcite{you2021LogME} & 0.021 & 0.075 & 0.627 & 0.417 & 0.146 & 0.743 & 0.763 & 0.686 & 0.738 & 0.260 & 0.181 & 0.423 \\
NLEEP \shortcite{li2021Nleep}& -0.286 & 0.662 & 0.595 & 0.108 & 0.374 & 0.779 & 0.598 & 0.716 & \textbf{0.864} & \underline{0.880} & -0.091 & 0.473 \\

SFDA \shortcite{shao2022SFDA} & \textbf{0.167} &0.674 & 0.683 & \textbf{0.846} & 0.789 & \underline{0.882} & \textbf{0.897} & 0.837 & 0.564 & 0.831 & \underline{0.621} & 0.708 \\
NCTI \shortcite{wang2023NCTI} & 0.036 & \textbf{0.811} & \textbf{0.796} & 0.758 & \underline{0.811} & 0.796 & 0.762 & \textbf{0.945} & \underline{0.805} & 0.774 & 0.606 & \underline{0.719} \\
    \midrule
DISCO & \underline{0.063} & \underline{0.691} & \underline{0.768} & \underline{0.825} & \textbf{0.835} & \textbf{0.918} & \underline{0.886} & \underline{0.898} & 0.542 & \textbf{0.916} & \textbf{0.642} & \textbf{0.726} \\
    \bottomrule
  \end{tabular}
  \caption{Method comparison of their correlation strength with ground-truth fine-tuning accuracy on self-supervised models. All the settings and notations are the same as Table \ref{table1}. Our method achieves the best overall weighted Kendall’s $\tau_\omega$. }
  \label{table2}
\end{table*}

\subsection{Hard-example Selection}
The naive implementation of singular value decomposition on extracted features $\mathbf{Z}\in \mathbb{R}^{N \times d}$ has a computational complexity of $\mathcal{O}(N^3d)$, which becomes impractical as $N$ increases. To address this issue, we propose a hard-example selection method to sample a subset of the datasets, thereby lowering 
$N$ and the overall time complexity. 
The method leverages the insight that a pre-trained model's performance on hard examples better reflects its transferability. We first project the features into a class-separable space using Linear Discriminant Analysis (LDA) with a complexity of $\mathcal{O}(Nd^2)$ when $N>d$, and then select the $N'$ data points with the lowest confidence scores from its classifier as hard examples (detailed in Appendix \ref{app:hardexample}).
This approach reduces time complexity, without causing a significant drop in performance. In practice, one can consider the trade-off between time and performance when choosing the sampling ratio.

\section{Experiments}

\subsection{Experimental Setup}
\paragraph{Evaluation Protocol}
To quantify the correlation between estimated assessment scores $\{S^m\}_{m=1}^M$ and actual fine-tuning results $\{P^m\}_{m=1}^M$, we use the weighted Kendall's $\tau_\omega$ (see Appendix \ref{app:tau}), which measures ranking agreement with higher weights for higher ranks. A larger $\tau_\omega$ indicates a more effective metric for ranking models. We supplement this with the top-$k$ probability, $Pr($top-k$)$, which measures the probability that the best ground-truth model is among the top $k$ estimated models. While $\tau_\omega$ evaluates overall ranking agreement, $Pr($top-k$)$ is particularly useful for identifying the best pre-trained model in practical scenarios.
\paragraph{Feature Construction}
We compute our DISCO over the features extracted by pre-trained models on the target dataset. For image classification, we perform once forward propagation to obtain the extracted feature $\mathbf{Z}\in \mathbb{R}^{N \times d}$. For object detection, varying object counts and additional bounding boxes complicate feature construction. We apply adaptive average pooling to unify the dimensions of box-specific features to a consistent size $\hat{h}$. The final feature matrix, $\mathbf{Z} \in \mathbb{R}^{K \times \hat{h}}$, is created by concatenating all normalized box-specific features, where $K$ is the total number of bounding boxes in the target dataset (detailed in Appendix \ref{app:feature}). 

\subsection{Image Classification}
\paragraph{Datasets}
We adopt 11 widely-used datasets in classification tasks, including FGVC Aircraft \cite{maji2013Aircraft}, Standford Cars \cite{krause2013StanfordCars}, Food101 \cite{bossard2014Food101}, Oxford-IIIT Pets \cite{parkhi2012Pets}, Oxford-102 Flowers \cite{nilsback2008Flowers}, Caltech101 \cite{fei2004Caltech}, CIFAR-10 \cite{krizhevsky2009CIFAR}, CIFAR-100 \cite{krizhevsky2009CIFAR}, VOC2007 \cite{everingham2010VOC2007}, SUN397 \cite{xiao2010SUN397}, DTD \cite{cimpoi2014DTD}. 
The fine-tuning performances (see Appendix \ref{app:fineres}) of both supervised CNN and self-supervised CNN models are obtained from \cite{shao2022SFDA}.



\subsubsection{Results and Analysis on Supervised Model}
\label{supervised}
\paragraph{Pre-trained Models}
We construct a pool of 11 widely-used models including ResNet-34 \cite{he2016ResNet}, ResNet-50 \cite{he2016ResNet}, ResNet-101 \cite{he2016ResNet}, ResNet-152 \cite{he2016ResNet}, DenseNet-121 \cite{huang2017DenseNet}, DenseNet-169
 \cite{huang2017DenseNet}, DenseNet-201 \cite{huang2017DenseNet}, MNet-A1 \cite{tan2019mnasnet}, MobileNet-v2 \cite{sandler2018mobilenetv2}, GoogleNet \cite{szegedy2015GoogleNet} and Inception-v3 \cite{szegedy2016Inceptionv3}. All these supervised source models were pre-trained on ImageNet and were downloaded from the PyTorch repository.

\paragraph{Performance Comparison}

We evaluate DISCO against established model selection metrics, including NCE \cite{tran2019NCE}, LEEP \cite{nguyen2020LEEP}, LogME \cite{you2021LogME}, SFDA \cite{shao2022SFDA},  NLEEP \cite{li2021Nleep}, ETran \cite{gholami2023etran} and NCTI \cite{wang2023NCTI}. Results in Table \ref{table1} show that DISCO achieves SOTA with an average $\tau_\omega$ of 0.739, a $1.8\%$ relative improvement over the second-best NCTI.
DISCO's superior performance is attributed to its ability to consider the fine-grained contribution of different spectral components during the fine-tuning process. 
This leads to strong results on fine-grained classification datasets like Aircraft, Cars, and Flowers, where the relative improvements are $31.5\%$, $22.9\%$, and $31.6\%$ compared to the NCTI, respectively. 

\begin{figure*}[ht]
    \centering
    \begin{subfigure}[t]{0.23\textwidth}
        \centering
        \includegraphics[width=\textwidth]{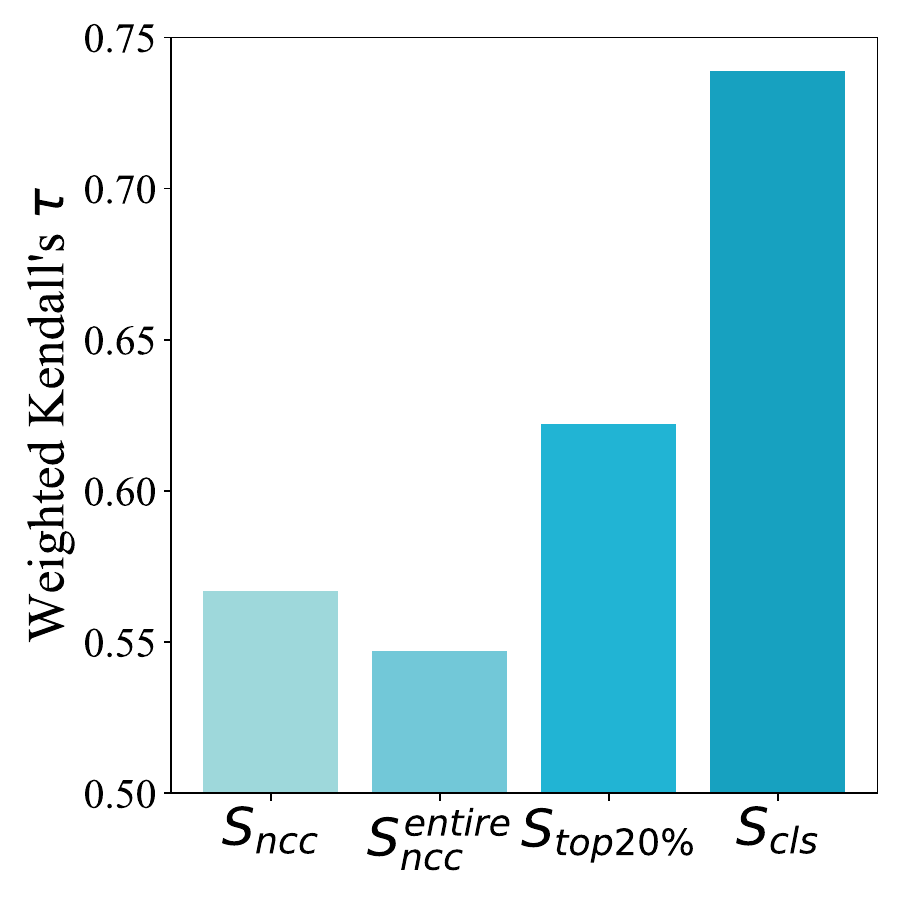}
        \caption{Framework ablation}
        \label{fig:supervised}
    \end{subfigure}
    \begin{subfigure}[t]{0.23\textwidth}
        \centering
        \includegraphics[width=\textwidth]{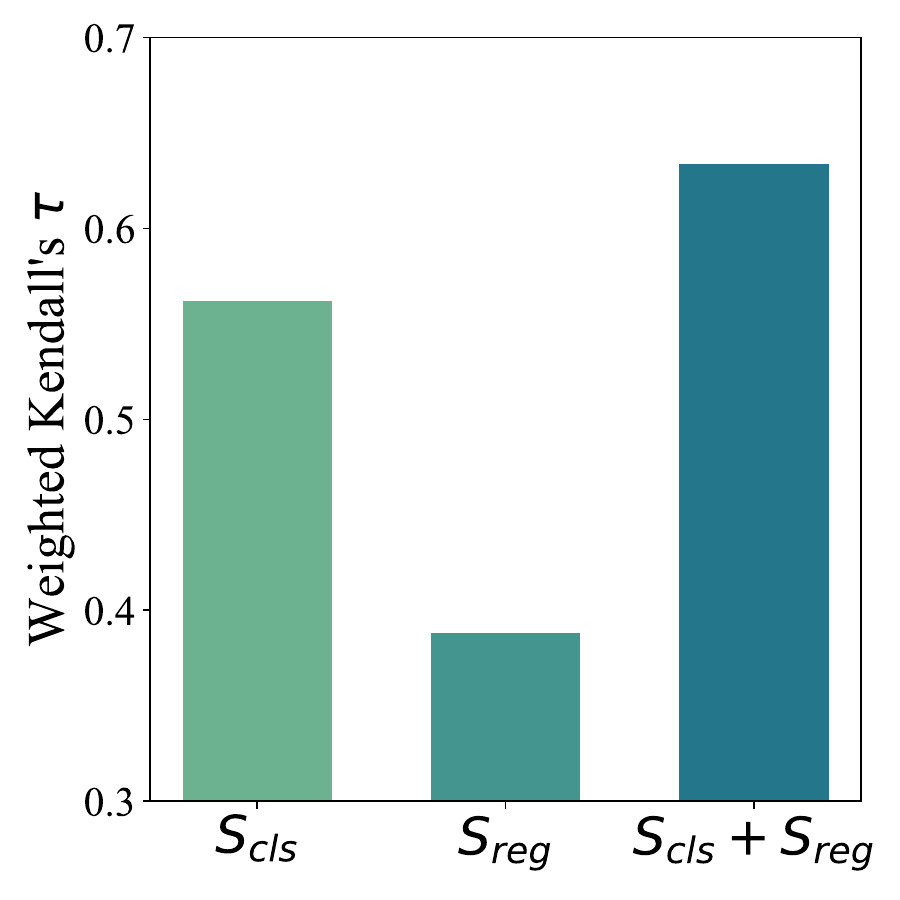}
        \caption{Object detection ablation}
        \label{fig:obj}
    \end{subfigure}
        \begin{subfigure}[t]{0.23\textwidth}
        \centering
        \includegraphics[width=\textwidth]{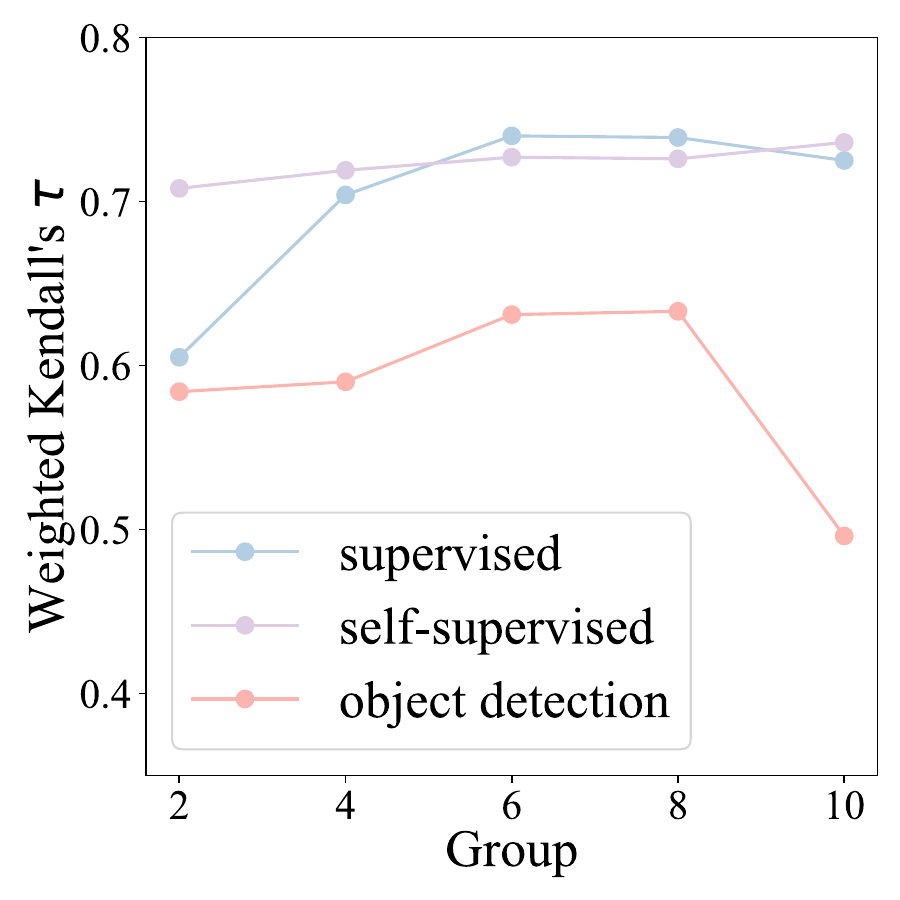}
        \caption{Choices of group numbers}
        \label{fig:compare}
    \end{subfigure}
    \begin{subfigure}[t]{0.23\textwidth}
        \centering
        \includegraphics[width=\textwidth]{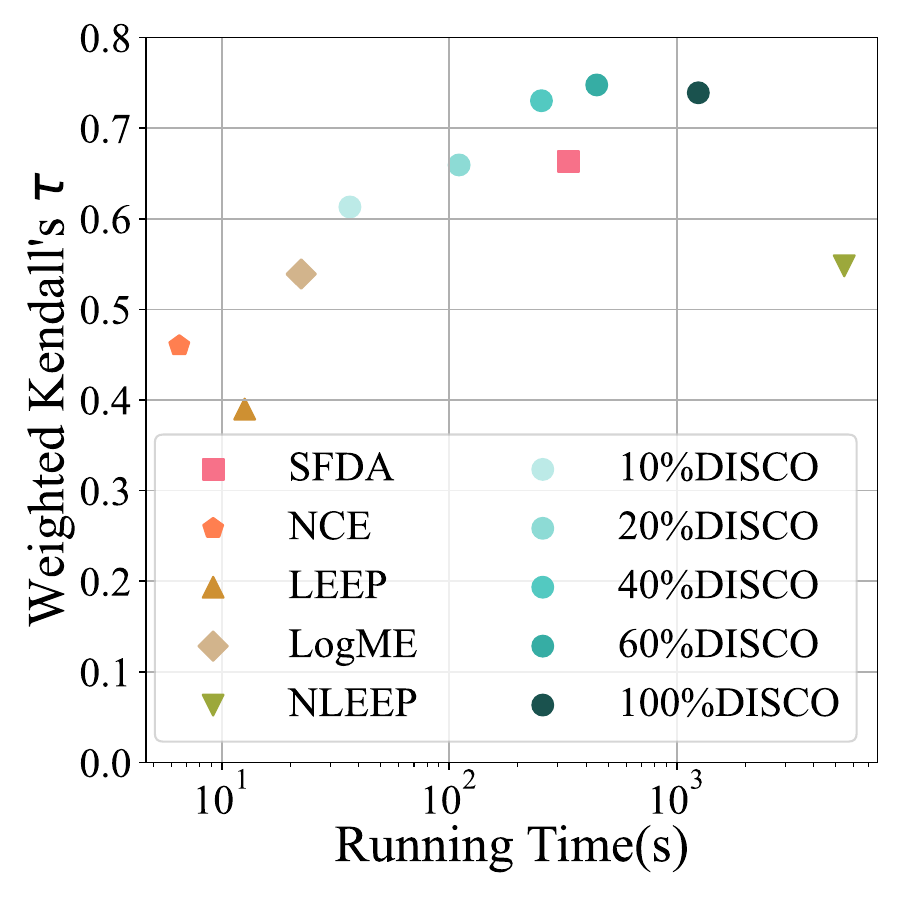}
        \caption{Time complexity analysis}
        \label{fig:time}
    \end{subfigure}
    \caption{(a) Ablation study on the framework and (b) on the object detection benchmark. (c) The average $\tau_\omega$ with different group numbers on three benchmarks. (d) Method comparison w.r.t average running time (seconds) and $\tau_\omega$ on 11 datasets.}
    \label{fig:ablation}
\end{figure*}

\subsubsection{Results and Analysis on Self-Supervised Model}
\label{self-supervised}
\paragraph{Pre-trained Models} 
To further assess the effectiveness of our metric in self-supervised learning (SSL), we select 10 SSL models based on ResNet-50 architecture \cite{he2016ResNet}, including BYOL \cite{grill2020BYOL}, Deepclusterv2 \cite{caron2018deepclusterv2}, Infomin \cite{tian2020Infomin}, MoCo-v1 \cite{he2020mocov1}, MoCo-v2 \cite{chen2020mocov2}, Instance Discrimination \cite{wu2018InsDis}, PCL-v1 \cite{li2020PCL}, PCL-v2 \cite{li2020PCL}, Sela-v2 \cite{asano2019selav2} and SWAV \cite{caron2020SWAV}.

\paragraph{Performance Comparison} 
Since NCE and LEEP require pre-trained classifiers, they are no longer suitable for ranking self-supervised models, which typically lack classifiers. We compare DISCO with NLEEP, LogME, SFDA and NCTI, as shown in Table \ref{table2}. DISCO performs consistently well in self-supervised model selection, while conventional metrics collectively tend to exhibit lower correlation scores. The DISCO leads and secures the highest average weighted Kendall's $\tau_\omega$ of 0.726 ranking correlation among 11 datasets. 
This superior performance highlights DISCO's ability to capture intricate details from spectral components, unlike traditional metrics that rely on overall feature characteristics. By leveraging this rich information, DISCO provides more effective model assessment for transfer learning.

\subsection{Object Detection}
\paragraph{Datasets}
We select five datasets that span different domains and sizes for object detection to evaluate our model selection metric, including Blood \cite{blood-cell-detection-1ekwu_dataset}, ForkLift \cite{forklift-dsitv_dataset}, NFL \cite{nfl-competition_dataset}, Valorant Video Game \cite{valorant-9ufcp_dataset}, and CSGO Video Game \cite{wlots_datasetcsgo}. 
\begin{table}[ht]
\centering

  \setlength{\tabcolsep}{1mm} 
    \begin{tabular}{lccccc}
    \toprule
    Method &  Pr(top1) & Pr(top2) & Pr(top3) & $\tau_\omega$ \\
    \midrule
    LogME \shortcite{you2021LogME}  &  0.600 & 0.600 & 0.800 & 0.374 \\
    PACTran \shortcite{ding2022pactran} &   0.400 & 0.400 & 0.600 & 0.140 \\
    Linear \shortcite{ding2022pactran}  &  0.200 & 0.400 & 0.600 & 0.214 \\
    SFDA \shortcite{shao2022SFDA}   & 0.400 & 0.600 & \textbf{1.000} & 0.312 \\
    \midrule
    LogME+$S_\text{lmr}$ & 0.600 & 0.600 & 0.800 & 0.400 \\
    PACtran+$S_\text{lmr}$ & 0.200 & 0.400 & 0.800 & 0.322 \\
    Linear+ $S_\text{lmr}$ & 0.200 & 0.200 & 0.800 & 0.306 \\
    SFDA+$S_\text{lmr}$ & 0.200 & 0.400 & 0.800 & 0.202 \\
    ETran \shortcite{gholami2023etran}  & 0.600 & 0.800& \textbf{1.000} & 0.522 \\
    \midrule
    DISCO & \textbf{1.000} & \textbf{1.000} & \textbf{1.000} & \textbf{0.634} \\
    \bottomrule
    \end{tabular}
    \caption{Results on object detection benchmark. Weighted Kendall's $\tau_\omega$ is the average of five object detection datasets.}
\label{tab:HF_results}
\end{table}

\paragraph{Pre-trained Models} 
To demonstrate the generality of DISCO for pre-trained models on object detection, we assemble a pool of six models: YOLOv5s, YOLOv5m, YOLOv5n \cite{jocher2023yolov5}, YOLOv8s, YOLOv8m, and YOLOv8n \cite{jocher2023yolo}, all pre-trained on the COCO dataset \cite{lin2014microsoftcoco}. We obtain the ground-truth fine-tuned results for these source models from the HuggingFace \cite{wolf2019huggingface}.

\paragraph{Performance Comparison} 


 We compare our DISCO with naive LogME, SFDA, PACTran \cite{ding2022pactran}, and Linear method \cite{ding2022pactran} using classification-only score, as well as their variants incorporating the regression score $S_{\text{lmr}}$ baseline proposed in Etran. The experimental comparison results of $\tau_\omega$ and $Pr($top-k$)$ are presented in Table \ref{tab:HF_results}. Generally, variants with the additional regression score baseline perform better than the naive estimation scores and Etran surpasses the regression baseline. However, our DISCO achieves the SOTA result of average weighted Kendall $\tau_\omega$ of 0.634, reflecting a $21\%$ improvement over Etran's $\tau_\omega$ of 0.522. The superior results show that our method is not only applicable in classification tasks but also excels in object detection, indicating the generality of our approach across a broad range of applications.


\subsection{Further Analysis}
\paragraph{Ablation Study}
Figure \ref{fig:supervised} shows the ablation results of $S_{\text{ncc}}$ ($\sum_{g=1}^G S_{\text{ncc}}^g$, w/o $S_{\text{ratio}}^g$ compared to our framawork), $S^{\text{entire}}_{\text{ncc}}$ ($S_\text{ncc}$ when $G=1$), $S_{\text{top20\%}}$ (top20\% singular values ratio) compared to our proposed $S_{\text{cls}}$ for image classification. These results highlight the effectiveness of DISCO in accounting for different spectral components with different contributions.
Figure \ref{fig:obj} presents the ablation results for each term in object detection, where their combination achieves the highest $\tau_\omega$, demonstrating the indispensability of the regression score in object detection and the flexibility of our framework. More details are in Appendix \ref{app:ablation}.

\paragraph{Choices of Group Numbers}
The number of groups for spectral components is a crucial factor in determining the performance of our method. As shown in Figure \ref{fig:compare}, small groups fail to differentiate between spectral components' contributions, while large groups cause each group to have too little information, reducing accuracy in downstream tasks. We found that $G=6$ or $G=8$ offer the best balance and are ideal for practical use.


\paragraph{Time Complexity}
We compare the running time vs. average $\tau_\omega$ for 11 datasets on supervised models using different model selection metrics in Figure \ref{fig:time}. DISCO with the full dataset delivers the best $\tau_\omega$ but has high time costs, albeit still far below fine-tuning. Using hard-example selection to sample 60\%, 40\%, 20\% and 10\% of the dataset reduces time complexity with minimal performance loss. Notably, at 20\% sampling, DISCO matches SFDA's performance with lower time costs, and at 10\%, it outperforms LogME with a similar time cost, demonstrating the efficiency of DISCO and the hard-example selection method. Detailed analyses are provided in Appendix \ref{app:timecomp}.

\section{Conclusion}
In this paper, we investigate the fine-grained changes in extracted features during fine-tuning and propose an assessment framework for pre-trained models based on the Distribution of Spectral Components (DISCO). We design tailored metrics for classification and regression, which successfully apply to image classification and object detection. Extensive experiments demonstrate that our method achieves SOTA in assessing pre-trained models for transfer learning.

\section*{Acknowledgments}
This work was supported by the National Natural Science Foundation of China (62406112, 62306085, 62372179) and Shenzhen College Stability Support Plan (GXWD20231130151329002).

\bibliography{aaai25}

\appendix
\newpage
\section{More Details about DISCO}
\label{sec:appA}
\subsection{Interpretation of Weighted Kendall’s tau}
\label{app:tau}
Kendall's $\tau$ is a statistical measure that evaluates the ordinal relationship between two sets of measured quantities, reflecting the degree of correspondence between their rankings. Given the estimated assessment scores $\{S^m\}_{m=1}^M$ and actual fine-tuning results $\{P^m\}_{m=1}^M$, Kendall's $\tau$ is calculated as:
\begin{equation}
    \tau = \frac{2}{M(M-1)}\sum_{1\leq i< j \leq M} sgn(P^i - P^j)sgn(S^i-S^j)
\end{equation}
where $\text{sgn}(x)$ is the sign function. 
Furthermore, we adopt a weighted version of Kendall's $\tau$, denoted as $\tau_\omega$, to quantify this correlation, which assigns higher weights to higher ranks. A larger $\tau_\omega$ indicates that the pre-trained model selection metric is more effective in ranking models. 
\subsection{Feature Construction}
\label{app:feature}



Unlike image classification, where feature construction is straightforward and achieved through once-forward propagation, object detection presents additional complexities due to varying object counts and bounding boxes. Directly utilizing the original forward-propagated features as inputs can lead to misalignment with the actual bounding boxes.
To address this, for the $i$-th image in the target dataset, which contains $B$ object detection boxes, the relative position of each $b$-th box is determined by its annotated coordinates. The box-specific feature, denoted by $f_i^{b}$, is then extracted based on its relative location within the original features.
Given the varying sizes of detection boxes, the height and width of the corresponding extracted features will differ. To unify these features into a consistent size, adaptive average pooling is applied, normalizing the dimensions to a fixed size $\hat{h}$. The final feature matrix, $\textbf{Z} \in \mathbb{R}^{K \times \hat{h}}$, is constructed by concatenating all the normalized box-specific features, where $K$ represents the total number of bounding boxes across the entire dataset.

Additionally, following established model selection methods \cite{bolya2021PARC,wang2023NCTI, pandy2022GBC}, we employ principal component analysis (PCA) to reduce the dimensions of the extracted features to a uniform 128 in our experiments, ensuring fair comparisons between different pre-trained models. This dimensionality reduction also reduces the noise and redundancy in the data, enhancing the robustness and reliability of our model selection approach.
\subsection{Hard-example Selection}
\label{app:hardexample}
In this section, we present the details of our hard-example selection method.
Given the extracted feature $\mathbf{Z}\in \mathbb{R}^{N \times d}$, we compute the projection matrix $\mathbf{W}$ using Linear Discriminant Analysis (LDA) as follows: 
\begin{equation} 
    \mathbf{W} =\mathop{\arg\max}\limits_{\mathbf{W}} \frac{\mathbf{W}^\mathbf{T} \mathbf{S_b} \mathbf{W}}{\mathbf{W}^\mathbf{T} \mathbf{S_w} \mathbf{W}}
\end{equation}
where $\mathbf{S_b} = \sum_{c=1}^C N_c (\bm{\mu}_c -\bm{\mu} )(\bm{\mu}_c -\bm{\mu} )^\mathsf{T}$ and $\mathbf{S_w}= \sum_{c=1}^C \sum_{n=1}^{N_c} (\mathbf{z}_n^{(c)} -\bm{\mu}_c )(\mathbf{z}_n^{(c)} -\bm{\mu}_c )^\mathsf{T}$ represent the between-class and within-class scatter matrices. Here, $N_c$ is the number of samples in class $c$, $\bm{\mu} = \sum_{n=1}^N \mathbf{z}_n$ and $\bm{\mu}_c = \sum_{n=1}^{N_c}\mathbf{z}_n^{(c)}$ denote the mean of the entire featrue and the mean of the $c$-th class, respectively.
Once the projection matrix $\mathbf{W}$ is obtained, we can project the original feature $\mathbf{Z}$ into a more class-separable space by $\hat{\mathbf{Z}} = \mathbf{W}^\mathsf{T} \mathbf{Z}$. For each class, we assume $\hat{\mathbf{Z}}^{(c)} \sim \mathcal{N}(\mathbf{W}^\mathsf{T}\bm{\mu}_c, \mathbf{\Lambda_c}$) where $\mathbf{\Lambda_c}$ represents the covariance matrix of $\hat{\mathbf{Z}}^{(c)}$. For simplicity, the linear version of LDA assumes $\mathbf{\Lambda_c} = \pmb{I}$, where $\pmb{I}$ is the identity matrix.
By applying Bayes theorem, given a sample $x_n$ with corresponding feature $\hat{\mathbf{z}}_n$, we can obtain the posterior probability $\delta_n^c$ that it belongs to the $c$-th class as follows:
\begin{equation}
\delta_n^c = \hat{\mathbf{z}}_n^\mathsf{T}\mathbf{W}\mathbf{W}^\mathsf{T}\bm{\mu}_c-\frac{1}{2}\bm{\mu}_c^\mathsf{T}\mathbf{W}\mathbf{W}^\mathsf{T}\bm{\mu}_c + \log(\pi_c)
\end{equation}
The prior probability $\pi_c$ is given by $N_c/N$ for class $c$.
We then apply the softmax function to normalize the posterior probability for each sample $\hat{\mathbf{z}}_n$:
\begin{equation}
    \hat{y}_{n,c} = \frac{\text{exp}(\delta_n^c)}{\sum_{k=1}^C\text{exp}(\delta_n^k)}
\end{equation}
where $\hat{y}_{n,c}$ is the final predicted probability that it belongs to class $c$. We define $\mathbf{\hat{y}}_n$ as a $C$-dimensional logits vector, with each dimension representing one class and the $y$-th position indicating the confidence that the sample belongs to its ground-truth class.
Then, for each class, we sort the $N_c$ samples by their confidence in ascending order and select the samples with the top$k$ lowest confidence scores as hard examples. 
This approach reduces the time complexity of our DISCO by lowering $N$.

\begin{table}[b]
\centering
 \setlength{\tabcolsep}{1mm} 
    \begin{tabular}{ccccc}
    \toprule
    Method & Pr(top1) & Pr(top2) & Pr(top3) & $\tau_\omega$ \\
    \midrule
    DISCO ($S_{\text{cls}}$) & 0.600 & 0.800 & 0.800 & 0.562 \\
    DISCO ($S_{\text{reg}}$) & 0.600 & 0.800 & \textbf{1.000} & 0.388 \\
    DISCO ($S_{\text{cls}}+S_{\text{reg}}$) & \textbf{1.000} & \textbf{1.000} & \textbf{1.000} & \textbf{0.634} \\
    \bottomrule
    \end{tabular}
\caption{Ablation study for object detection.}
\label{tab:HF_ablation}
\end{table}

\begin{figure*}[tp]
    \centering
    \includegraphics[width=\textwidth]{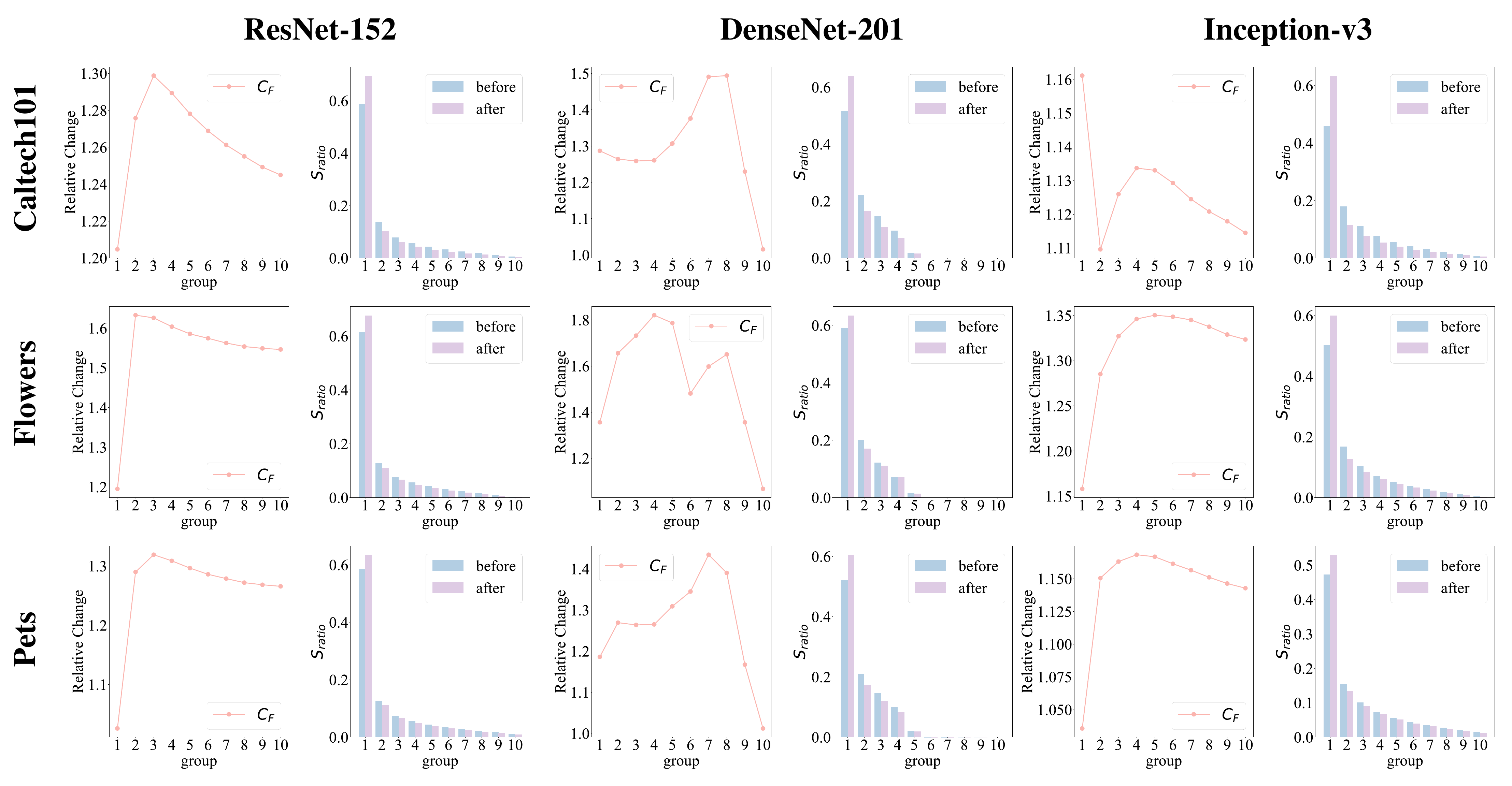}
    \caption{The relative changes of Frobenius norm $C_F$ and the proportion of singular values $S_{\text{ratio}}$ in different spectral components of extracted features before and after fine-tuning.}
    \label{fig:completeres}
\end{figure*}

\begin{table*}[bp]

  \setlength{\tabcolsep}{1mm} 
  \centering
  \begin{tabular}{c|ccccccccccccc}
    \toprule
dataset & &  Aircraft &Caltech &Cars &Cifar10 &Cifar100 &DTD &Flowers &Food &Pets &SUN &VOC &Average \\
\hline
\multirow{2}{*}{20\%   random} 
& $\tau_\omega$ & -0.057  & 0.540  & 0.788 & 0.823 & 0.890 & 0.479 & 0.740 & 0.738 & 0.578 & 0.708 & 0.629 & 0.623  \\
& time(s) & 23.3  & 18.9  & 48.8 & 98.9 & 182.0 & 9.4 & 17.2 & 315.5 & 7.5 & 181.2 & 6.3 & 82.6  \\
\multirow{2}{*}{20\% DISCO}  
& $\tau_\omega$  & 0.619 & 0.489 & 0.556 & 0.872 & 0.552  & 0.808 & 0.613  & 0.495  & 0.736 & 0.643 & 0.870 & 0.659 \\
& time(s) & 50.4 & 40.8 & 78.6  & 121.8  & 215.3 & 13.8 & 28.5 & 348.8 & 10.3  & 298.5 & 8.2  & 110.5 \\
\hline
\multirow{2}{*}{40\%   random} 
& $\tau_\omega$  & 0.219 & 0.587 & 0.439 & 0.823 & 0.895  & 0.649 & 0.491  & 0.678 & 0.649 & 0.739 & 0.693 & 0.624  \\
& time(s) & 40.3  & 20.8 & 68.1 & 323.8 & 537.9 & 12.6 & 17.6 & 986.5  & 8.9  & 323.2 & 7.4 & 213.4 \\
\multirow{2}{*}{40\% DISCO}    
& $\tau_\omega$ & 0.573 & 0.618 & 0.867 & 0.894 & 0.925 & 0.608 & 0.720 & 0.595 & 0.603 & 0.770 & 0.860 & 0.730 \\
& time(s) & 67.8 & 29.4 & 83.5 & 350.7 & 606.5  & 16.4 & 24.5 & 1111.4 & 13.1 & 484.3 & 9.6 & 254.3  \\
\hline
\multirow{2}{*}{60\%   random} 
& $\tau_\omega$  & 0.205 & 0.600 & 0.516 & 0.844 & 0.863 & 0.835 & 0.611 & 0.678 & 0.378 & 0.739 & 0.678 & 0.631 \\
& time(s) & 52.6 & 26.7 & 103.7 & 689.2 & 915.1 & 18.5 & 19.8 & 1938.3 & 13.8 & 495.5 & 10.9 & 389.5 \\
\multirow{2}{*}{60\% DISCO}    
& $\tau_\omega$  & 0.551 & 0.566 & 0.835 & 0.910 & 0.900 & 0.759 & 0.786 & 0.647 & 0.679 & 0.721 & 0.775 & \textbf{0.739} \\
& time(s) & 73.3 & 33.6 & 106.8 & 714.0 & 1265.4 & 15.1 & 24.7 & 2126.7 & 14.7 & 508.0 & 13.7 & 445.1 \\
\hline
\multirow{2}{*}{80\%   random} 
& $\tau_\omega$  & 0.509 & 0.680 & 0.804 & 0.823 & 0.863  & 0.685 & 0.657 & 0.657 & 0.671 & 0.773 & 0.814 & 0.721 \\
& time(s) & 110.8 & 43.6 & 182.6 & 1715.9 & 2016.8  & 26.1  & 35.9 & 4907.7 & 24.6 & 899.3 & 29.6  & 908.4  \\
\multirow{2}{*}{80\% DISCO}    
& $\tau_\omega$  & 0.628 & 0.646 & 0.827 & 0.910 & 0.938 & 0.447 & 0.746 & 0.680 & 0.714 & 0.781 & 0.817 & \textbf{0.739} \\
& time(s) & 97.1  & 37.9  & 155.8 & 1282.3 & 1748.9 & 18.1 & 32.8 & 3880.0 & 21.3 & 732.8 & 20.4 & 729.8  \\
\hline
\multirow{2}{*}{100\% DISCO} 
& $\tau_\omega$ & 0.652 & 0.661 & 0.795 & 0.823 & 0.895 & 0.764& 0.712 & 0.678 & 0.575 & 0.773 & 0.802 & \textbf{0.739} \\
& time(s) & 161.9 & 82.8 & 189.5 & 3038.4 & 3206.6 & 30.4 & 34.5 & 5756.6 & 30.6 & 1112.7 & 51.9 & 1245.1 \\      
\bottomrule
  \end{tabular}
    \caption{Compared results of hard-example selection method to the random sampling method on supervised models. Weighted Kendall's $\tau_\omega$ and time cost on 11 target datasets and their average are listed above. Hard-example selection method achieves a higher average weighted Kendall’s $\tau_\omega$ than the random sampling method. }
  \label{tab:timecomp}
\end{table*}

\begin{table*}[ht]
\centering
\begin{tabular}{lcccccccccccc}
\toprule
 & Aircraft & Caltech & Cars & Cifar10 & Cifar100 & DTD & Flowers & Food & Pets & SUN & VOC \\ 
 \midrule
ResNet-34 & 84.06 & 91.15 & 88.63 & 96.12 & 81.94 & 72.96 & 95.2 & 81.99 & 93.5 & 61.02 & 84.6 \\
ResNet-50 & 84.64 & 91.98 & 89.09 & 96.28 & 82.8 & 74.72 & 96.26 & 84.45 & 93.88 & 63.54 & 85.8 \\
ResNet-101 & 85.53 & 92.38 & 89.47 & 97.39 & 84.88 & 74.8 & 96.53 & 85.58 & 93.92 & 63.76 & 85.68 \\
ResNet-152 & 86.29 & 93.1 & 89.88 & 97.53 & 85.6 & 76.44 & 96.86 & 86.28 & 94.42 & 64.82 & 86.32 \\
DenseNet-121 & 84.66 & 91.5 & 89.34 & 96.45 & 82.75 & 74.18 & 97.02 & 84.99 & 93.07 & 63.26 & 85.28 \\
DenseNet-169 & 84.19 & 92.51 & 89.02 & 96.77 & 84.26 & 74.72 & 97.32 & 85.84 & 93.62 & 64.1 & 85.77 \\
DenseNet-201 & 85.38 & 93.14 & 89.44 & 97.02 & 84.88 & 76.04 & 97.1 & 86.71 & 94.03 & 64.57 & 85.67 \\
MNet-A1 & 66.48 & 89.34 & 72.58 & 92.59 & 72.04 & 70.12 & 95.39 & 71.35 & 91.08 & 56.56 & 81.06 \\
MobileNetV2 & 79.68 & 88.64 & 86.44 & 94.74 & 78.11 & 71.72 & 96.2 & 81.12 & 91.28 & 60.29 & 82.8 \\
Googlenet & 80.32 & 90.85 & 87.76 & 95.54 & 79.84 & 72.53 & 95.76 & 79.3 & 91.38 & 59.89 & 82.58 \\
InceptionV3 & 80.15 & 92.75 & 87.74 & 96.18 & 81.49 & 72.85 & 95.73 & 81.76 & 92.14 & 59.98 & 83.84 \\ 
\bottomrule
\end{tabular}
\caption{The fine-tuning accuracy of 11 supervised models on 11 target tasks.}
\label{fineres:sup}
\end{table*}

\begin{table*}[ht]
\centering
\begin{tabular}{lcccccccccccc}
\toprule
 & Aircraft & Caltech & Cars & Cifar10 & Cifar100 & DTD & Flowers & Food & Pets & SUN & VOC \\ 
 \midrule
BYOL & 82.1 & 91.9 & 89.83 & 96.98 & 83.86 & 76.37 & 96.8 & 85.44 & 91.48 & 63.69 & 85.13 \\
Deepclusterv2 & 82.43 & 91.16 & 90.16 & 97.17 & 84.84 & 77.31 & 97.05 & 87.24 & 90.89 & 66.54 & 85.38 \\
Infomin & 83.78 & 80.86 & 86.9 & 96.72 & 70.89 & 73.47 & 95.81 & 78.82 & 90.92 & 57.67 & 81.41 \\
InsDis & 79.7 & 77.21 & 80.21 & 93.08 & 69.08 & 66.4 & 93.63 & 76.47 & 84.58 & 51.62 & 76.33 \\
MoCov1 & 81.85 & 79.68 & 82.19 & 94.15 & 71.23 & 67.36 & 94.32 & 77.21 & 85.26 & 53.83 & 77.94 \\
MoCov2 & 83.7 & 82.76 & 85.55 & 96.48 & 71.27 & 72.56 & 95.12 & 77.15 & 89.06 & 56.28 & 78.32 \\
PCLv1 & 82.16 & 88.6 & 87.15 & 96.42 & 79.44 & 73.28 & 95.62 & 77.7 & 88.93 & 58.36 & 81.91 \\
PCLv2 & 83.0 & 87.52 & 85.56 & 96.55 & 79.84 & 69.3 & 95.87 & 80.29 & 88.72 & 58.82 & 81.85 \\
Sela-v2 & 85.42 & 90.53 & 89.85 & 96.85 & 84.36 & 76.03 & 96.22 & 86.37 & 89.61 & 65.74 & 85.52 \\
SWAV & 83.04 & 89.49 & 89.81 & 96.81 & 83.78 & 76.68 & 97.11 & 87.22 & 90.59 & 66.1 & 85.06 \\ 
\bottomrule
\end{tabular}
\caption{The fine-tuning accuracy of 10 self-supervised models on 11 target tasks.}
\label{fineres:self}
\end{table*}

\section{More Experimental Results}
\label{sec:appB}
\subsection{Empirical Experiments of Fine-tuning}
\label{app:entireemp}
The fine-tuning process of our empirical experiments follows the setting in \cite{shao2022SFDA}. We represent the complete results of our empirical experiments, showing the changes in different spectral components of the extracted features before and after fine-tuning in Figure \ref{fig:completeres}. 

Each row represents the same dataset, and each column represents the same model. For each model on each dataset, we present the relative changes in the Frobenius norm $C_F$ (left) and the proportion of singular values $S_{\text{ratio}}$ (right) for different spectral components before and after fine-tuning.

\subsection{More Details of Ablation Study}
\label{app:ablation}
We represent the detailed ablation results of our object detection benchmark. The contributions of each term to the overall score on object detection are shown in Table \ref{tab:HF_ablation}. Among the two individual terms of DISCO, it is evident that $S_{\text{cls}}$ achieves a higher $\tau_\omega$ while $S_{\text{reg}}$ is more accurate in predicting the top$k$ pre-trained models. The combination of them achieves the highest $\tau_\omega$ and $Pr($topk$)$ for pre-trained models. This result demonstrates the indispensability of the regression score in object detection and the flexibility of our framework.

\begin{table}
\centering
\setlength{\tabcolsep}{1mm} 
\begin{tabular}{lccccc}
\toprule
 & NFL & Blood & CSGO & Forklift & Valorant \\ 
 \midrule
Yolov5s & 0.261 & 0.902 & 0.924 & 0.838 & 0.982 \\
Yolov5m & 0.314 & 0.905 & 0.932 &0.852 & 0.990\\
Yolov5n & 0.217 & 0.923 & 0.908 & 0.789 & 0.959 \\
Yolov8s & 0.279 & 0.917 & 0.886 & 0.851 & 0.971 \\
Yolov8m & 0.287 & 0.927 & 0.892 & 0.846 & 0.965 \\
Yolov8n & 0.209 & 0.893 & 0.844 & 0.838 & 0.937 \\ 
\bottomrule
\end{tabular}
\caption{The fine-tuning accuracy (map50) of pre-trained models on object detection benchmark.}
\label{fineres:object}
\end{table}

\subsection{More Time Complexity Analysis}

\label{app:timecomp}
We further compared our hard-example selection method to the random sampling method on the supervised model benchmark. The compared results are represented in Table \ref{tab:timecomp}. Selecting 20\%, 40\%, 60\%, and 80\% of the dataset using hard-example selection reduces time complexity while maintaining better performance than the random sampling method. Using $40\%$ of the data, DISCO achieves an average $\tau_\omega$ of 0.730 on the supervised model benchmark, better than SFDA's 0.663, while keeping average time costs per dataset at 254.3 seconds, well below SFDA's 333.3 seconds.
Besides, SOTA methods such as SFDA mainly sacrifice some efficiency for accuracy. Compared with them, DISCO is both more accurate and efficient, achieving a better trade-off between these two sides. We believe that DISCO’s slightly higher time cost is worthwhile for the benefit of improved accuracy. 
Furthermore, as a model selection score, DISCO’s computational cost is already far lower than that of a brute-force fine-tuning process. For example, DISCO takes around 280 seconds per model on $100\%$ of CIFAR10 and 32 seconds per model on $40\%$ of CIFAR10, whereas fine-tuning each model on CIFAR10 takes about 16,000 seconds on the same GPU. This makes DISCO's time cost quite acceptable for model selection.

\begin{table*}[ht]

\centering
\setlength{\tabcolsep}{1mm} 
\begin{tabular}{lcccccccccccc}
\toprule
 & Aircraft & Caltech & Cars & Cifar10 & Cifar100 & DTD & Flowers & Food & Pets & SUN & VOC & Average\\ 
 \midrule

LogME &\textbf{0.297}&0.141&0.311&0.284&0.196&0.257&0.035&0.099&-0.150&-0.163&0.239&0.141 \\ 
SFDA &0.231&0.049&0.386&0.406&0.164&0.300&0.134&0.251&0.471&0.345&0.366&0.282\\
DISCO&0.187&\textbf{0.700}&\textbf{0.582}&\textbf{0.556}&\textbf{0.601}&\textbf{0.624}&\textbf{0.566}&\textbf{0.620}&\textbf{0.482}&\textbf{0.615}&\textbf{0.557}&\textbf{0.554}\\
\bottomrule
\end{tabular}
\caption{Method comparison of their correlation strength with ground-truth fine-tuning accuracy on the extended benchmark. All the settings and notations are the same as Table 1. Our method achieves the best overall weighted Kendall’s $\tau_\omega$.}
\label{extended}
\end{table*}

\begin{table*}[ht]

\centering
\setlength{\tabcolsep}{1mm} 
\begin{tabular}{lcccccccccccc}
\toprule
 & Aircraft & Caltech & Cars & Cifar10 & Cifar100 & DTD & Flowers & Food & Pets & SUN & VOC & Average\\ 
 \midrule

LogME &\textbf{0.450}&0.345&0.503&0.575&0.202&0.348&0.150&0.109&0.222&0.227&0.588&0.338 \\ 
SFDA &0.435&0.267&0.468&0.644&0.300&0.336&0.284&0.154&0.606&0.396&0.329&0.383\\
DISCO&0.074&\textbf{0.576}&\textbf{0.710}&\textbf{0.785}&\textbf{0.883}&\textbf{0.790}&\textbf{0.771}&\textbf{0.695}&\textbf{0.751}&\textbf{0.798}&\textbf{0.785}&\textbf{0.693}\\
\bottomrule
\end{tabular}
\caption{Method comparison of their correlation strength with ground-truth fine-tuning accuracy on the supervised models. We consider models with accuracy differences within $0.1\%$ as having the same rank and recalculate $\tau_\omega$. Our method achieves the best overall weighted Kendall’s $\tau_\omega$.}
\label{recalculation}
\end{table*}

\subsection{Results of ground truth of Fine-tuning}
\label{app:fineres}
The ground truth of pre-trained model ranking is obtained by fine-tuning all pre-trained models with a hyper-parameters sweep on target datasets. The detailed ground truth of fine-tuning for supervised models, self-supervised models and the object detection benchmark are shown in Table \ref{fineres:sup}, Table \ref{fineres:self}, Table \ref{fineres:object}, respectively.

\subsection{Extended Benchmark}

To investigate the performance of our method when using pre-trained models from diverse architectures and different pre-training paradigms, we developed an extended benchmark comprising 24 pre-trained models: 11 supervised CNN models, 10 self-supervised CNN models (already described in our paper), and an additional 3 Vision Transformers (ViT-S, ViT-B, and Swin-T). Given that NCE and LEEP require pre-trained classifiers, these methods are no longer suitable for ranking self-supervised models. Therefore, we compare our DISCO to existing methods, SFDA and LogME. The weighted Kendall's $\tau_\omega$ on 11 target datasets, along with their average, are listed in Table \ref{extended}.
The results clearly show that our DISCO method performs consistently well across a broader range of pre-trained models. This highlights the robustness and reliability of our proposed method.

\subsection{Rethinking of Correlation Calculation}

To evaluate the correlation between our estimated assessment scores and actual fine-tuning results, we use weighted Kendall's $\tau_\omega$ following the settings in \cite{shao2022SFDA, gholami2023etran, wang2023NCTI}. However, upon examining the accuracies obtained by different models after fine-tuning (Tables 6 and 7), we noticed that several models exhibit practically identical performance. This observation suggests that the small differences in accuracy might not be statistically significant and could introduce noise into the evaluation of the proposed metrics. 
To mitigate this issue, we consider models with accuracy differences within $0.1\%$ as having the same rank and recalculate $\tau_\omega$. This adjustment reduces the impact of negligible differences on the ranking, ensuring a more robust evaluation. The recalculated weighted Kendall's $\tau_\omega$ for 11 target datasets on supervised models, along with their average, are summarized in Table \ref{recalculation}. 
Even with this adjustment, DISCO outperforms SFDA and LogME, achieving average $\tau_\omega$ of 0.693, 0.383 and 0.338 respectively, which underscores the robustness and reliability of our proposed method.

\end{document}